\documentclass[journal]{IEEEtran}
\usepackage{algorithm}
\usepackage{algorithmic}
\usepackage{graphicx}
\usepackage{stfloats}
\usepackage{subfigure}
\usepackage{float}
\usepackage{booktabs}
\usepackage{threeparttable}
\usepackage{multirow}
\usepackage{longtable}

\usepackage{amsmath}
\usepackage{amsfonts}
\usepackage{mathrsfs}
\usepackage{amssymb}
\usepackage{amsthm}

\usepackage{array}
\ifCLASSINFOpdf
\else
   \usepackage[dvips]{graphicx}
\fi
\usepackage{url}

\usepackage{graphicx}

\begin{document}

\title{Local Neighbor Propagation Embedding}

\author{Shenglan~Liu,~\IEEEmembership{Member,~IEEE},
  Yang~Yu
  \IEEEcompsocitemizethanks{\IEEEcompsocthanksitem S. Liu, Y. Yu are with the School of Computer Science and Technology, Faculty of Electronic Information and Electrical Engineering, Dalian University of Technology, Dalian 116024, Liaoning, China. E-mail: \{liusl, yyu\}@mail.dlut.edu.cn}%

  \thanks{Manuscript received April 19, 2005; revised August 26, 2015.\protect\\
    (Corresponding author: S. Liu.)\protect\\}
}

\markboth{Journal of \LaTeX\ Class Files,~Vol.~14, No.~8, August~2015}%
{Shell \MakeLowercase{\textit{et al.}}: Bare Demo of IEEEtran.cls for IEEE Journals}

\maketitle

\begin{abstract}
	Manifold Learning occupies a vital role in the field of nonlinear dimensionality reduction and its ideas also serve for other relevant methods. Graph-based methods such as Graph Convolutional Networks (GCN) show ideas in common with manifold learning, although they belong to different fields. Inspired by GCN, we introduce neighbor propagation into LLE and propose Local Neighbor Propagation Embedding (LNPE). With linear computational complexity increase compared with LLE, LNPE enhances the local connections and interactions between neighborhoods by extending $1$-hop neighbors into $n$-hop neighbors. The experimental results show that LNPE could obtain more faithful and robust embeddings with better topological and geometrical properties.
\end{abstract}

\begin{IEEEkeywords}
	Manifold Learning, Local Neighbor Propagation, Topological and Geometrical Properties
\end{IEEEkeywords}

\IEEEpeerreviewmaketitle

\section{Introduction}
\IEEEPARstart{O}{ver} the past few decades, manifold learning has already caused broad attention and applied in biological science\cite{becht2019dimensionality, moon2018manifold}, image reconstruction \cite{zhu2018image}, pose estimation \cite{raytchev2004head, benabdelkader2010robust, wang2017head}, etc. The history of manifold learning can be tracked to some local algorithms such as Locally Linear Embedding (LLE) algorithm \cite{roweis2000nonlinear}, Local Tangent Space Alignment (LTSA) algorithm \cite{zhang2003nonlinear} and some global algorithms such as the Isometric Mapping (ISOMAP) algorithm \cite{tenenbaum2000global} and the Maximum Variance Unfolding (MVU) algorithm \cite{weinberger2006introduction}. Besides, corresponding to nonlinear manifold learning methods, there are some linear embedding methods which are more conductive to practical applications such as Locality Preserving Projection (LPP) \cite{he2004locality}, Neighborhood Preserving Embedding (NPE) \cite{he2005neighborhood}, Neighborhood Preserving Projection (NPP) \cite{kokiopoulou2005orthogonal}. However, it’s difficult for linear methods to deal with complex nonlinear data. In recent years, manifold learning has been used in many applications such as text \cite{salem2017manifold}, image \cite{wang2018flexible, pedronette2018unsupervised, pedronette2016correlation}, audio \cite{arevalo2018manifold} and video \cite{wang2018flexible, yang2016online}.

Locally linear methods in manifold learning such as LLE has been widely proposed and applied. Hessian Locally Linear Embedding (HLLE) \cite{donoho2003hessian}, Modified Locally Linear Embedding (MLLE) \cite{zhang2007mlle} and Improved Locally Linear Embedding (ILLE) \cite{xiang2011regression} are some improved versions for LLE. Among the LLE-based methods, HLLE has high computational complexity and MLLE would not run on weak-connected or even worse data distributions. Actually, computational or algorithmic complexity and robustness are issues to be addressed for many improved LLE-based methods. This years, Graph Convolutional Network (GCN) becomes an important topic and graph-based algorithms reflect similar idea with manifold learning. In this paper, we improve LLE and propose a simple and unambiguous method named Local Neighbor Propagation Embedding (LNPE), which are introduced neighbor propagation used in GCN. LNPE extends the neighborhood size through neighbor propagation layer by layer and enhance the topological connections within each neighborhood. From the view of global mapping, neighborhood interactions increase with neighbor propagation to improve the global geometrical properties. With linear computational complexity increase, we propose a simple but effective framework through LNPE. Experimental results verify the effectiveness of our LNPE and performance with different dataset and neighborhood size show the robustness.

The remainder of this paper is organized as follows. We first introduce related work in Section 2. Section 3 shows the main body of this paper, which includes the motivation of LNPE, the mathematical background, the LNPE framework and computational complexity analysis. The experimental results are presented in Section 4 to verify the effectiveness and robustness of LNPE and Section 5 summarize our work in this paper.

\section{Related Work}
The essence of manifold learning is how to maintain the relationship corresponding to the intrinsic structure between samples in two different spaces. Researchers have done lots of work to measure the relationship from different aspects. 

PCA maximizes the global variance to reduce dimensionality, while Multidimensional Scaling (MDS) \cite{MDS} considers the low-dimensional distance between samples that is consistent with high-dimensional data. Based on MDS, ISOMAP utilize the shortest path algorithm to realize global mapping. Besides, MVU realizes an "unfolding" manifold through positive semi-definite and kernel technique, and RML \cite{RML} obtains the intrinsic structure of manifold with Riemannian method instead of Euclidean distance. Compared with ISOMAP, LLE represents local manifold learning, which obtains neighbor weights with locally linear reconstruction. Furthermore, Locally Linear Coordination (LLC) \cite{LLC} constructs a local model and makes a global alignment and contributes to both LLE and LTSA. LTSA describes local curvature through the tangent space of samples and takes the curvature as the weight of tangent space to realize global alignment. 

Another genre in manifold learning is graph-based embedding. Classical graph-based manifold learning methods include LE \cite{LE} and its linear version LPP. More related algorithms include NPE, Orthogonal Neighborhood Preserving Projections (ONPP) \cite{ONPP}, etc. Graph method produces a far-reaching influence on machine learning and related fields. For instance, graph is introduced into semi-supervised learning in LE, and LLE is also a kind of neighbor graph. Moreover, $L$-1 graph-based methods such as Sparsity preserving Projections (SPP) \cite{SPP} and its supervised extension \cite{gui2012discriminant} occurs with the wide application of sparse methods.

\section{Local Neighbor Propagation Embedding}
In this section, the idea of neighbor propagation is introduced gradually. The motivation of LNPE will be firstly stated to explain the origin of LNPE, and then we will give a review of the basic algorithm prototype. Finally, the LNPE framework is naturally expressed.
\subsection{Motivation}
LLE, which is a classically effective method, should be a representative of local information based methods in manifold learning. In the case of simple data distribution, LLE tends to get satisfactory results. But once the data distribution becomes complex or sparse, it is difficult for LLE to maintain topological and geometrical properties. The following items show the reasons.

\begin{enumerate}
	\item The neighborhood size is hard to determine for complex or sparse data. Inappropriate neighbors will be selected with a larger neighborhood size.
	\item LLE focuses more on each single neighborhood, but is weak in the interaction between different neighborhoods. Thus, it is difficult to obtain the ideal effect in geometrical structure preserving.
\end{enumerate}

There is a natural contradiction between Item 1 and Item 2. More specifically, the interaction of neighborhood information will be weakened by small neighborhood inevitably, which will produce unconvincing geometrical and topological structure of embedding results. In order to improve the capability and robustness of LLE, we introduce neighbor propagation into LLE and propose Local Neighbor Propagation Embedding (LNPE), inspired by GCN. The 1-hop neighborhood is extended to $n$-hop neighborhood through $n$ neighbor propagation, which will enhance the connections of points in different neighborhoods. In this way, LNPE avoids short circuit selection (Item 1) by setting a small neighborhood size $k$. Meanwhile, $i$-hop neighbors expands neighborhood size to depict local parts adequately by enhancing the topological connections. Furthermore, correlations between different neighborhoods (Item 2) are generated in neighbor propagation to produce more overlapping information, which is conducive to preserve geometrical structure.

\subsection{Mathematical Background}
Based on LLE, LNPE introduces neighbor propagation to improve the applicability. We firstly review the original LLE before the LNPE framework is stated. 

Suppose ${\bf X}=\{{\bf x}_1, \ldots, {\bf x}_n\}$ $\subset {\mathrm{R}}^{D}$ indicates a high-dimensional dataset which lies on a smooth $d$-dimensional manifold approximately, LLE tends to embed the intrinsic manifold from high-dimensional space into lower-dimensional subspace with preserving geometrical and topological structures. Based on the assumption of local linearity, LLE firstly reconstructs each high-dimensional data point ${\bf x}_i$ through linear combination within each neighborhood ${\bf N}_{i}$, where ${\bf N}_{i}$ indicates the $k$-nearest neighbors of ${\bf x}_i$ and ${\bf N}_{i}=\{{\bf x}_{i_1}, \ldots, {\bf x}_{i_k}\}$. Then the reconstruction weights matrix $\bf W$ in high-dimensional space can be determined by minimizing the total reconstruction error ${\varepsilon_1}$ of all data points. 

\begin{equation}
\label{LLE-high}
{\varepsilon_1 ({\bf W})} = {\left\| {\bf X} {\bf W} - {\bf X} \right\|_{F}^{2}},
\end{equation}

\noindent where ${\bf W} = [\vec{\bf w}_1, \vec{\bf w}_2, \cdots, \vec{\bf w}_n] \in \mathrm{R}^{n \times n}$, the $i$-th column vector $\vec{\bf w}_i$ indicates the reconstruction weights of data point ${\bf x}_{i}$ and $\| \cdot \|_F$ is the Frobenius norm. To remove the influence of transformations including translation, scaling, and rotation, a sum-to-one constraint $\vec{\bf w}_i^T \vec{1} = 1$ is enforced for each neighborhood. 

Let ${\bf Y} = \{{\bf y}_{1},\dots,{\bf y}_{n}\} \subset \mathrm{R}^{d}$ be the corresponding dataset in low dimensions. The purpose of LLE is to preserve the same local structures reconstructed in high-dimensional space. Then in the low-dimensional space, LLE chooses to utilize the same weights $\bf W$ to reproduce the local properties. The objective is to minimize the total cost function

\begin{equation}
\label{LLE-low}
{\varepsilon_2}({\bf Y}) = {\left\| {\bf Y} {\bf W} - {\bf Y} \right\|_{F}^{2}},
\end{equation}

\noindent under the constraint ${\bf YY}^T = {\bf I}$. Thus, the high-dimensional coordinates are finally mapped into lower-dimensional observation space.

\subsection{Local Neighbor Propagation Framework}
LLE would obtain unsatisfactory embedding results when facing complex data, where the local and global data properties are not easy to maintain. A faithful embedding is root in more sufficient within-neighborhood and between-neighborhood information of local and global structure. For LLE, the interactive relationship in single neighborhood is not enough to reproduce detailed data distribution, especially when the neighborhood size $k$ is not large enough. The neighborhood propagation is introduced into LLE to intensify the topological connections within neighborhoods and interactions between neighborhoods.

\textbf{High-dimensional reconstruction.} Based on the single reconstruction in each neighborhood, LNPE propagates neighborhoods and determines propagating weight matrix. Similarly, we define the high-dimensional and low-dimensional data as ${\bf X}=\{{\bf x}_1, \ldots, {\bf x}_n\} \subset {\mathrm{R}}^{D}$ and ${\bf Y} = \{{\bf y}_{1},\dots,{\bf y}_{n}\} \subset \mathrm{R}^{d}$, respectively. Suppose that we have finished the single reconstruction with LLE (Eq. \ref{LLE-high}) and got the weight matrix ${\bf W}_1$. In the meantime, the first reconstructed data ${\bf X}^{(1)} = {\bf X} {\bf W}_1$ are obtained from the first reconstruction with LLE. LNPE is to reuse the reconstructed data such as ${\bf X}_1$ to reconstruct the original data points in ${\bf X}$ again with the reconstructed data previously. Then the first neighborhood propagation is to utilize ${\bf X}_1$ to reconstruct original data ${\bf X}$ through

\begin{equation}
\label{LNPE-high-1}
{\varepsilon}({\bf W}_2) = {\left\| {\bf X} {\bf W}_{1} {\bf W}_{2} - {\bf X} \right\|_{F}^{2}},
\end{equation}

\noindent where ${\bf W}_2$ corresponds to the weight matrix in the first neighbor propagation. Thus, we can simply combine the weight matrices ${\bf W}_1$ and ${\bf W}_2$ as ${\bf W} = {\bf W}_1 {\bf W}_2$. Compared with ${\bf W}_1$ in LLE, ${\bf W}$ extends the $1$-hop neighbors to $2$-hop neighbors, which expands the neighborhood size through neighbor propagation. One of the most important advantages is that neighbor propagation could enhance topological connections while avoiding short circuit. Besides, a truth worthy of note is that the multi-hop neighbors hold lower weights in the process of propagation. Then after $i-1$ propagations, each data point to be reconstructed establishes relations with its $i$-hop neighbors. The ($i-1$)-th neighbor propagation can be formulated as 

\begin{equation}
\label{LNPE-high-n}
{\varepsilon}({\bf W}_i) = {\left\| {\bf X} {\bf W}_{1} \cdots {\bf W}_{i-1} {\bf W}_{i} - {\bf X} \right\|_{F}^{2}},
\end{equation}

\noindent where ${\bf W}_i$ indicates the weight matrix in the ($i-1$)-th neighbor propagation. From the perspective of weight solution and optimization, the $i$-th neighbor propagation depends on all the first $i-1$ weight matrices and each weight matrix in the propagation must be determined in turn. 

\textbf{Global low-dimensional mapping.} Similar to LLE, the low-dimensional embedding in LNPE is to reproduce the high-dimensional properties determined in the reconstruction. LNPE aims to preserve all the topological connections from $1$-hop neighborhoods to $n$-hop neighborhoods with $n$ weight matrices. We define that the matrix product ${\bf P}_i = {\bf W}_1 {\bf W}_2 \cdots {\bf W}_i$ is the product of weight matrices in the first $i-1$ neighbor propagation. And then the low-dimensional total optimization function can be expressed as

\begin{equation}
\label{LNPE-low-original}
{\varepsilon}({\bf Y}) = \sum_{i=1}^{t+1} {\left\| {\bf Y} {\bf P}_{i} - {\bf Y} \right\|_{F}^{2}},
\end{equation}

\noindent where the parameter $t+1$ denotes the total hop in the high-dimensional reconstruction with $t$ neighbor propagations. According to the properties of F-norm, the low-dimensional objective Eq. \ref{LNPE-low} can be written as 

\begin{equation}
\label{LNPE-low}
{\varepsilon}({\bf Y}) = {\bf Y} \left( \sum_{i=1}^{t+1} ({\bf P}_i - {\bf I}) ({\bf P}_i - {\bf I})^T \right) {\bf Y}^T,
\end{equation}

\noindent where ${\bf I} \in \mathrm{R}^{n\times n}$ indicates the identity matrix. Under the constraint ${\bf Y} {\bf Y} = I$, the low-dimensional coordinates can be easily obtained by decomposing the target matrix. It should be noted that the bottom eigenvalue of ${\bf M}$ is 0, so ${\bf Y}$ actually consists of the bottom $2 \sim d+1$ eigenvectors. The detailed algorithm is shown as Algorithm \ref{al-LNPE}.

\begin{algorithm}
	\caption{\textit{LNPE Algorithm}}
	\label{al-LNPE}
	\begin{algorithmic}[1]
	  \REQUIRE ~~\\
	  high-dimensional data ${\bf X} \subset \mathrm{R}^{D}$;\\ neighborhood size $k$;\\ target dimensionality $d$;\\
	  the neighbor propagation times $t$.
	  \ENSURE ~~\\
	  low-dimensional coordinates ${\bf Y} \subset \mathrm{R}^d$;
	  \STATE Find the $k$-nearest neighbors ${\bf N}_i = \{{\bf x}_{i_1},\cdots,{\bf x}_{i_k}\}$ for each data point.
	  \STATE Compute the weight matrix ${\bf W}_1$ with LLE.
	  \STATE Initialize a zero matrix ${\bf M}$.
	  \FOR{$e = 1 : t$}
	  \STATE Compute the matrix product ${\bf P}_e = {\bf W}_1 {\bf W}_2 \cdots {\bf W}_e$
	  \STATE Compute ${\bf M} = {\bf M} + ({\bf P}_i - {\bf I}) ({\bf P}_i - {\bf I})^T$
	  \STATE Compute the reconstruction data ${\bf X}^{(e)} = {\bf X} {\bf P}_e$
	  \FOR{each sample ${\bf x}_i,i=1,\cdots,n$}
	  \STATE Compute $\vec{\bf w}_i^{{(e+1)}}$ in ${\bf W}_{e+1}$ with ${\bf X}$ and ${\bf X}^{(e)}$ through minimizing $\| {\bf x}_i - {\bf X}^{(e)} \vec{\bf w}_i^{{(e+1)}} \|_2^2$
	  \ENDFOR
	  \ENDFOR
	  \STATE Compute ${\bf M} = {\bf M} + ({\bf P}_{t+1} - {\bf I}) ({\bf P}_{t+1} - {\bf I})^T$
	  \STATE Solve ${\bf Y} = \arg\ \min\limits_{\bf Y} {\rm Tr}({\bf Y} {\bf M} {\bf Y}^T)$
	\end{algorithmic}
\end{algorithm}

Eq. \ref{LNPE-low-original} and Eq. \ref{LNPE-low} show that LNPE aims to preserve all the learned properties in the high-dimensional reconstruction. Specifically speaking, for the sequence $i = 1, 2, \cdots, t$, a smaller $i$ aims at maintaining the topological connections within each neighborhood, while a larger $i$ pays more attention to expand the neighborhood interactions between different neighborhoods.

\begin{figure*}[!t]
	\centering
	\subfigure[]{
	  \includegraphics[height=3.7cm]{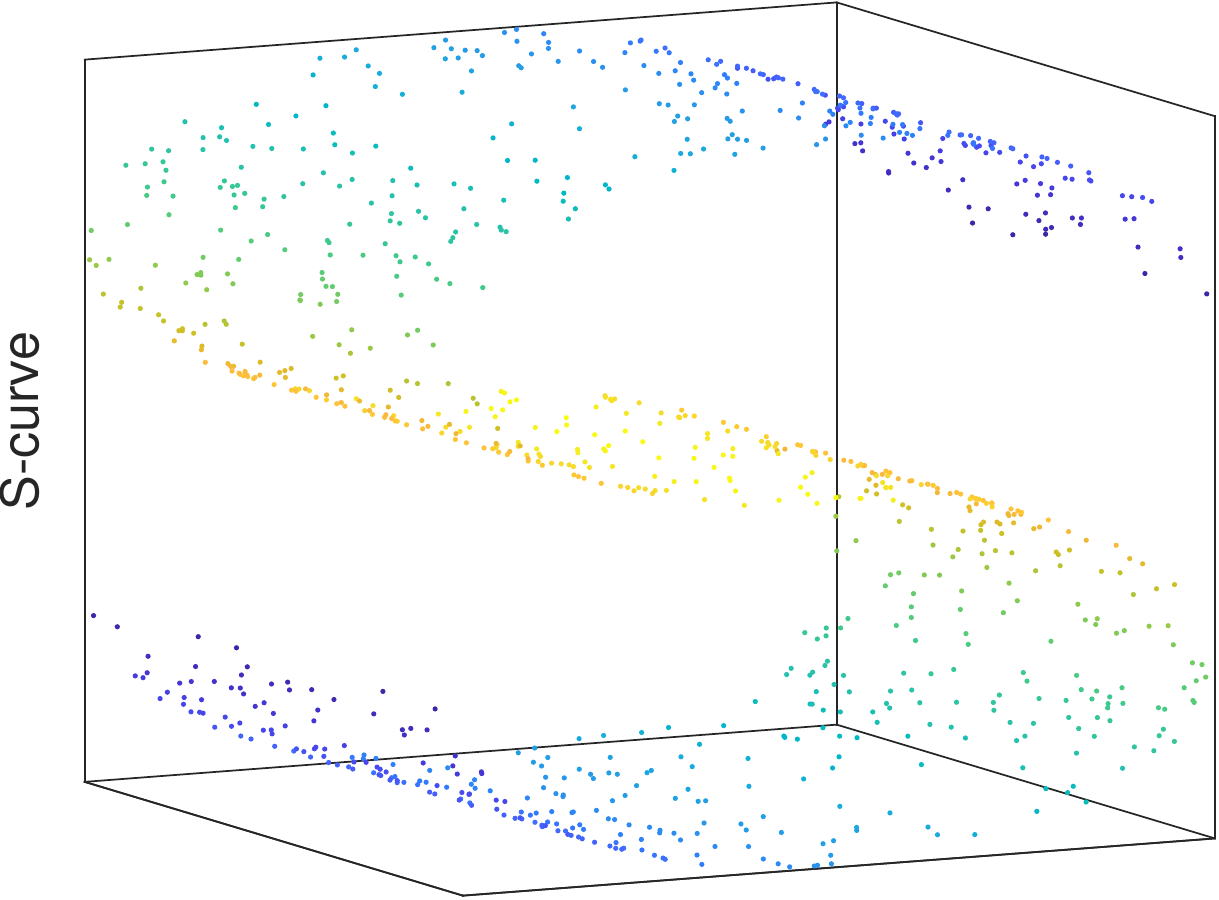}
	}
	\subfigure[]{
	  \includegraphics[height=3.7cm]{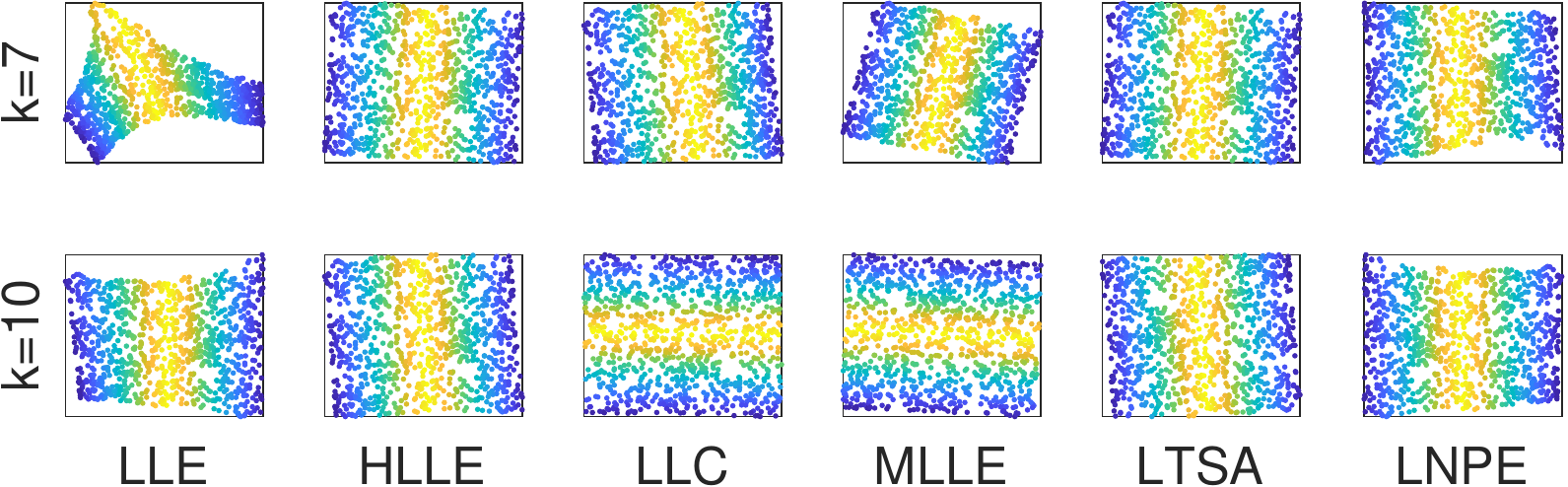}
	}
	\subfigure[]{
	  \includegraphics[height=3.7cm]{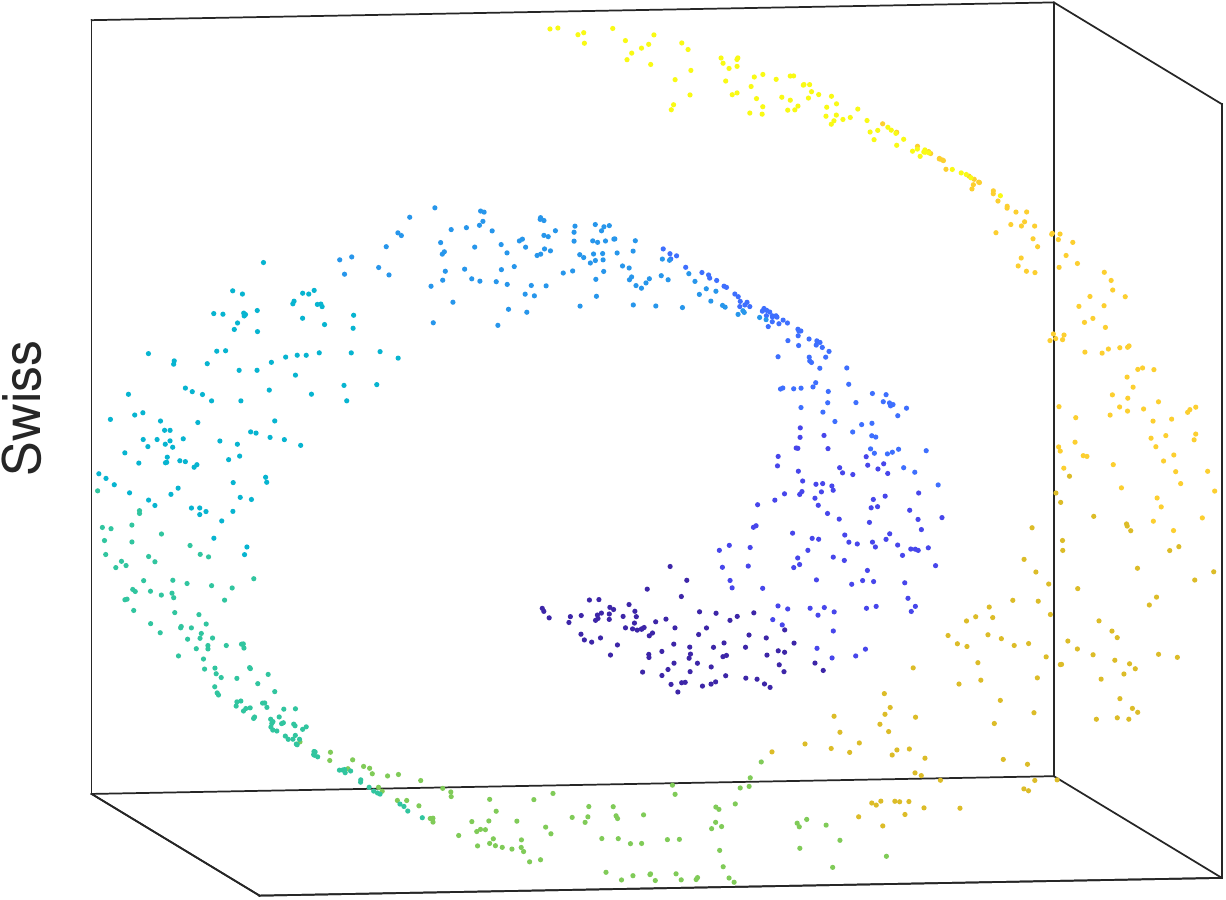}
	}
	\subfigure[]{
	  \includegraphics[height=3.7cm]{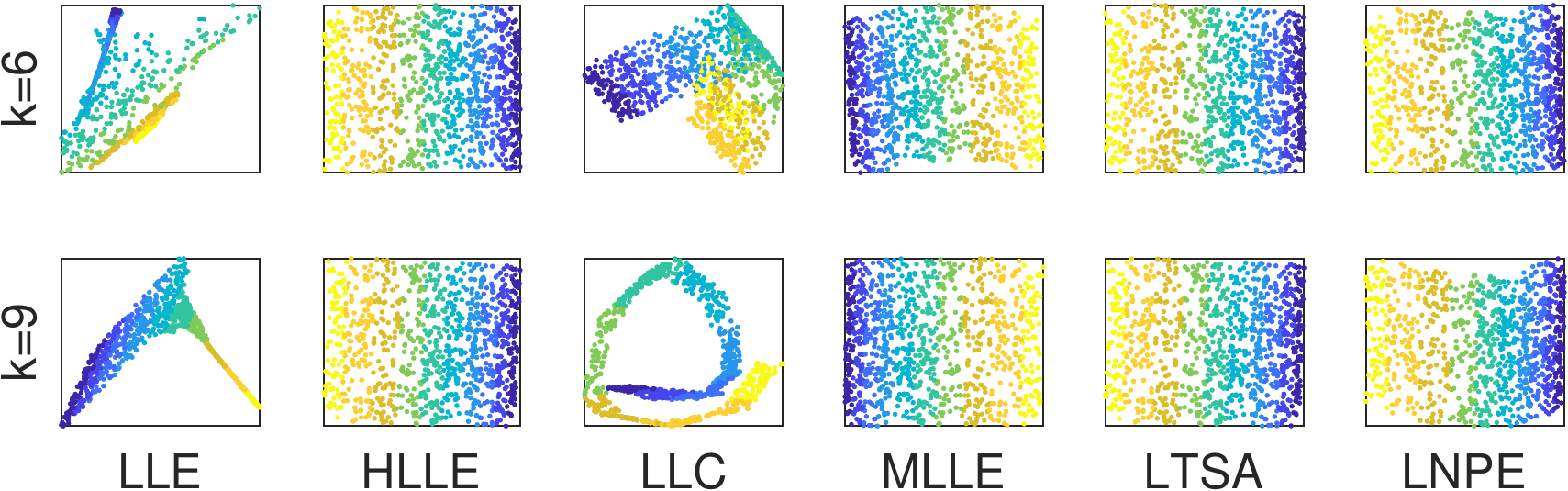}
	}
	\subfigure[]{
	  \includegraphics[height=3.7cm]{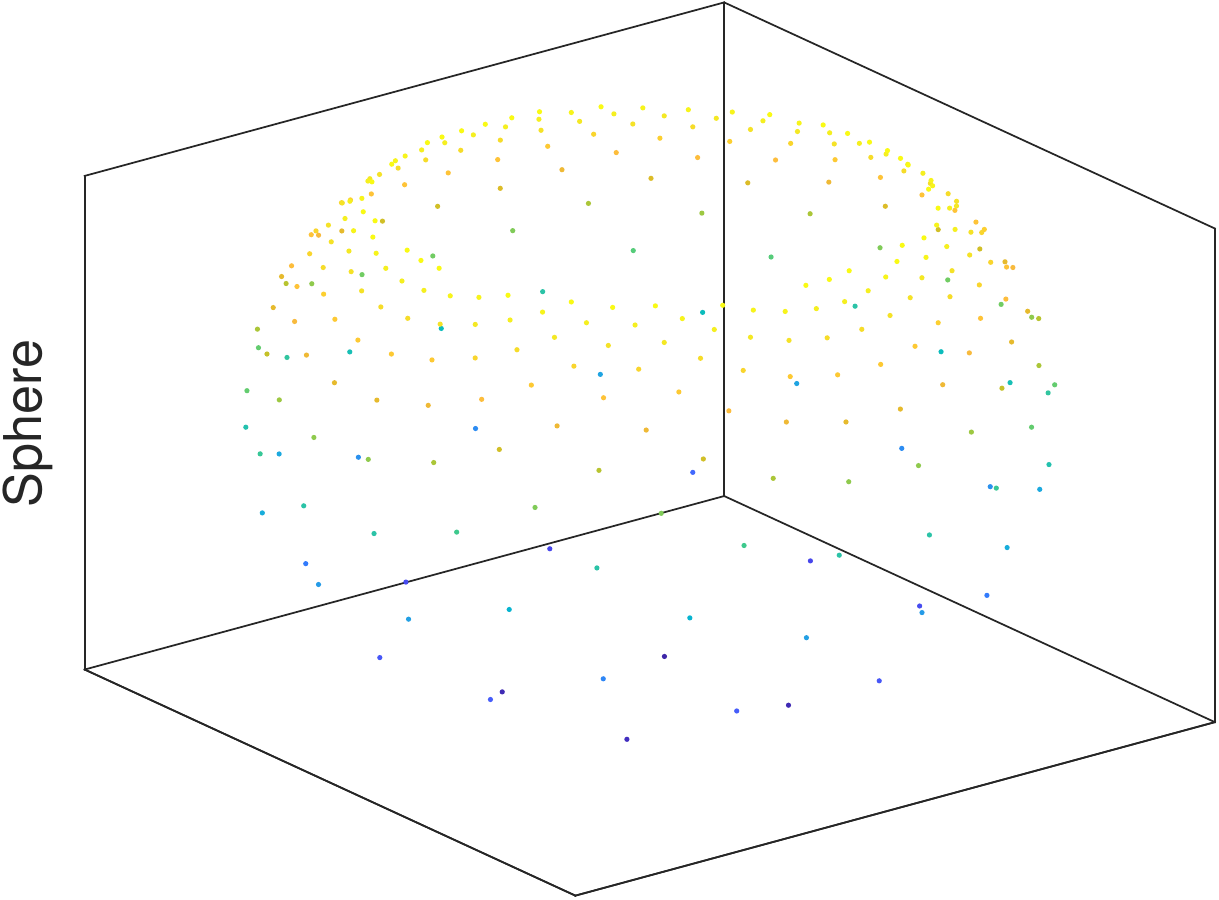}
	}
	\subfigure[]{
	  \includegraphics[height=3.7cm]{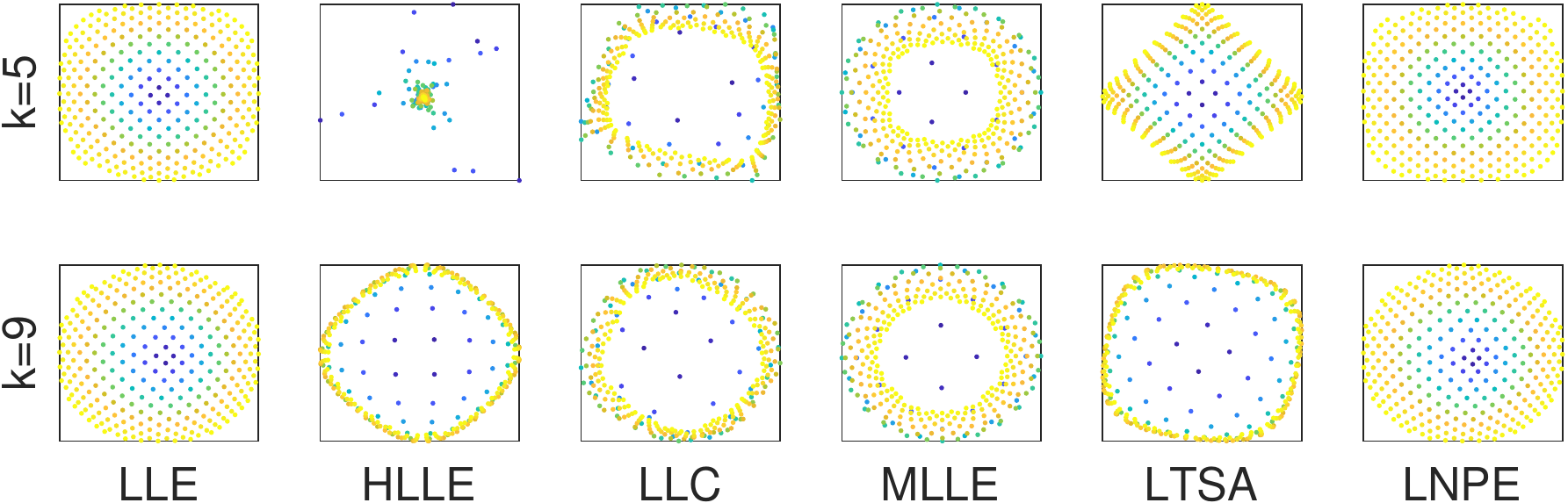}
	}
	\subfigure[]{
	  \includegraphics[height=3.7cm]{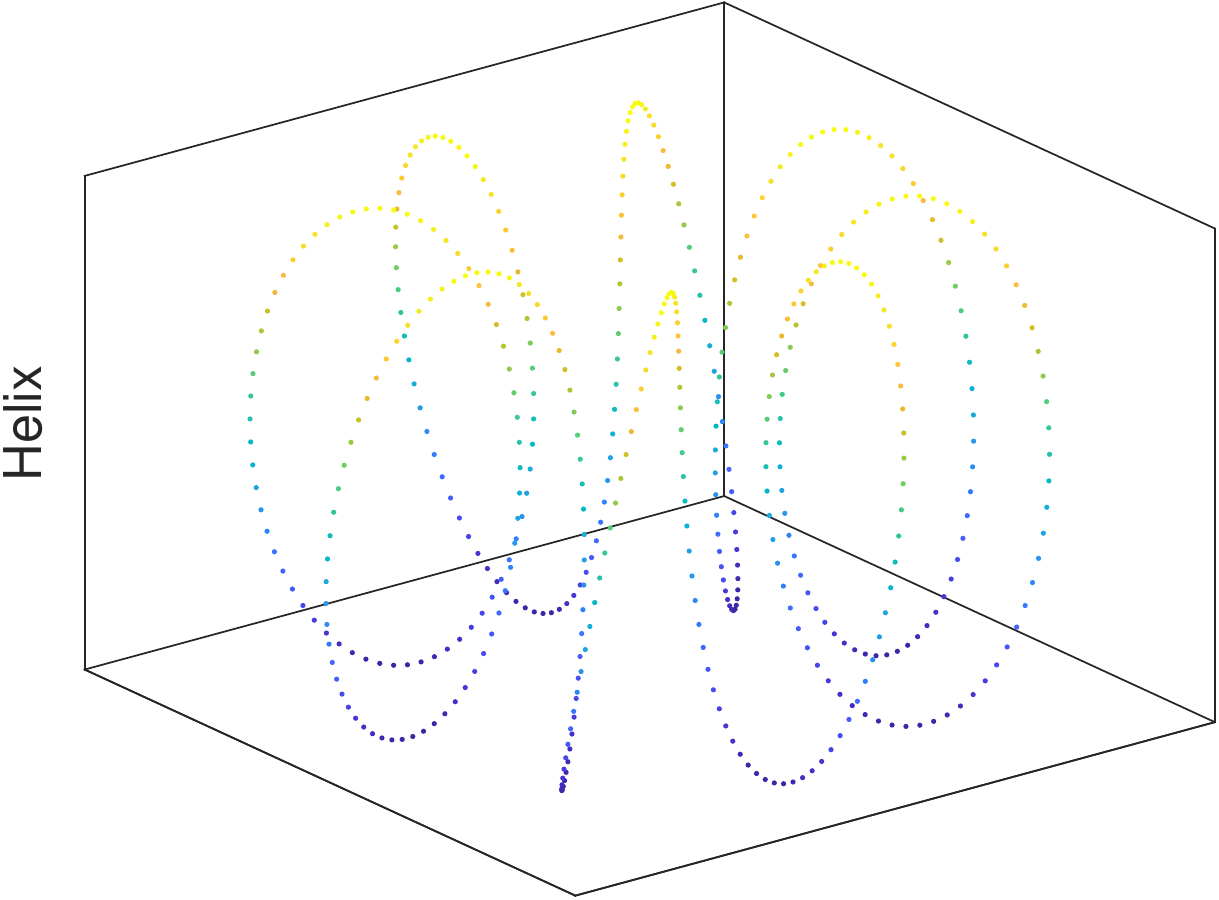}
	}
	\subfigure[]{
	  \includegraphics[height=3.7cm]{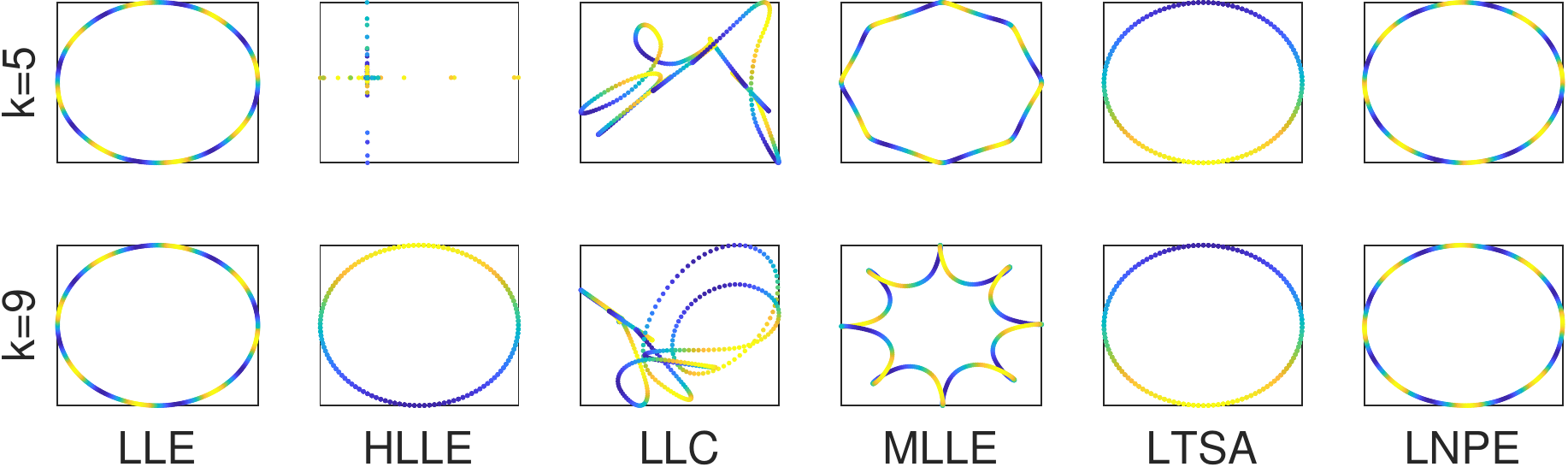}
	}
	\caption{Experiments with LLE, HLLE, LLC, MLLE, LTSA and LNPE on several synthetic datasets. (a) The S-Curve dataset with $n=1000$ samples. (b) The embedding results on S-Curve dataset. (c) The Swiss-Roll dataset with $n=1000$ samples. (d) The embedding results on Swiss-Roll dataset. (e) The Sphere dataset with $n=300$ samples. (f) The embedding results on Sphere dataset. (g) The Helix dataset with $n=500$ samples. (h) The embedding results on Helix dataset.}
	\label{exp-syn}
\end{figure*}

\subsection{Computational Complexity}
The computational complexity of LNPE naturally follows LLE. Calculating the $k$ nearest neighbors scales as $O(Dn^2)$. In some special data distributions, the computational complexity can be reduced to $O(n\log n)$ with K-D trees \cite{saul2000introduction}. Computing the weight matrix in $t+1$ reconstructions scales as $O((t+1)nk^3)$. Besides, computing matrices ${\bf P}$ and ${\bf M}$ scales as $O(p_1n^3)$ with sparse matrix and the final calculation of eigenvectors of a sparse matrix has computational complexity $O(p_2n^2)$, where $p_1$ and $p_2$ is parameters related to ratio of nonzero elements in sparse matrix \cite{van2009dimensionality}. 

\section{Experimental results}
Synthetic experiments are conducted in this section to verify the effectiveness of LNPE. We select 4 common synthetic dataset and limit the data amount $n$ between 1000 and 300. More specifically, we choose 5 LLE-related algorithms as the compared methods and each method is run on these 4 synthetic datasets to compare the dimensionality reduction results. For each dataset, we set 2 different neighborhood size $k$ to observe the changes of experimental results and verify the robustness of LNPE. For the  regularization coefficient $\sigma$ in LNPE and any compared methods that need a regularization in the listed experiments, it is set as $10^{-3}, 10^{-4}, 10^{-2}, 10^{-2}$ on S-Curve, Swiss-Roll, Sphere, Helix dataset, respectively.

Fig. \ref{exp-syn} shows the embedding results in the dimensionality reduction with LNPE and several compared methods. For S-Curve dataset in Fig. \ref{exp-syn} (b), LLE performs poorly when the neighborhood size $k$ is 7 and the embedding performance get improvements when $k$ is larger. Compared with LLE, HLLE, LLC, MLLE, LTSA and our LNPE all obtain faithful results. More specifically, through neighbor propagation, LNPE improves the topological and geometrical properties when $k$ is smaller. Besides, for Swiss-Roll dataset, because of the local high curvature and short circuit, LLE fails in preserving intrinsic manifold structure. But LNPE shows high performance, because neighbor propagation in LNPE enhances the local topological connections and neighborhood interactions. Furthermore, LLE and LNPE are better at addressing data such as Sphere and Helix dataset shown in Fig. \ref{exp-syn} (f) and (h). As $k$ changes from 5 to 9, LLE and LNPE could obtain stable results. It also demonstrates that LNPE, which introduce neighbor propagation into LLE, retains the feasible properties of LLE. 

\section{Conclusion}
LLE usually fail in addressing complex data where the local connections and neighborhood interactions are difficult to obtain. We introduce neighbor propagation in GCN into LLE and propose LNPE to improve the embedding performance. Compared with original LLE, LNPE could obtain more faithful embeddings with linear computational complexity increase. The experimental results show that our LNPE improves embedding performance and is more robust than LLE. 

\appendices

\section*{Acknowledgment}

Acknowledgment

\bibliography{test}

\begin{thebibliography}{10}
\providecommand{\url}[1]{#1}
\csname url@samestyle\endcsname
\providecommand{\newblock}{\relax}
\providecommand{\bibinfo}[2]{#2}
\providecommand{\BIBentrySTDinterwordspacing}{\spaceskip=0pt\relax}
\providecommand{\BIBentryALTinterwordstretchfactor}{4}
\providecommand{\BIBentryALTinterwordspacing}{\spaceskip=\fontdimen2\font plus
\BIBentryALTinterwordstretchfactor\fontdimen3\font minus
  \fontdimen4\font\relax}
\providecommand{\BIBforeignlanguage}[2]{{%
\expandafter\ifx\csname l@#1\endcsname\relax
\typeout{** WARNING: IEEEtran.bst: No hyphenation pattern has been}%
\typeout{** loaded for the language `#1'. Using the pattern for}%
\typeout{** the default language instead.}%
\else
\language=\csname l@#1\endcsname
\fi
#2}}
\providecommand{\BIBdecl}{\relax}
\BIBdecl

\bibitem{becht2019dimensionality}
E.~Becht, L.~McInnes, J.~Healy, C.-A. Dutertre, I.~W. Kwok, L.~G. Ng,
  F.~Ginhoux, and E.~W. Newell, ``Dimensionality reduction for visualizing
  single-cell data using umap,'' \emph{Nature biotechnology}, vol.~37, no.~1,
  p.~38, 2019.

\bibitem{moon2018manifold}
K.~R. Moon, J.~S. Stanley~III, D.~Burkhardt, D.~van Dijk, G.~Wolf, and
  S.~Krishnaswamy, ``Manifold learning-based methods for analyzing single-cell
  rna-sequencing data,'' \emph{Current Opinion in Systems Biology}, vol.~7, pp.
  36--46, 2018.

\bibitem{zhu2018image}
B.~Zhu, J.~Z. Liu, S.~F. Cauley, B.~R. Rosen, and M.~S. Rosen, ``Image
  reconstruction by domain-transform manifold learning,'' \emph{Nature}, vol.
  555, no. 7697, pp. 487--492, 2018.

\bibitem{raytchev2004head}
B.~Raytchev, I.~Yoda, and K.~Sakaue, ``Head pose estimation by nonlinear
  manifold learning,'' in \emph{Proceedings of the 17th International
  Conference on Pattern Recognition, 2004. ICPR 2004.}, vol.~4.\hskip 1em plus
  0.5em minus 0.4em\relax IEEE, 2004, pp. 462--466.

\bibitem{benabdelkader2010robust}
C.~BenAbdelkader, ``Robust head pose estimation using supervised manifold
  learning,'' in \emph{European conference on computer vision}.\hskip 1em plus
  0.5em minus 0.4em\relax Springer, 2010, pp. 518--531.

\bibitem{wang2017head}
C.~Wang, Y.~Guo, and X.~Song, ``Head pose estimation via manifold learning,''
  \emph{Manifolds-Current Research Areas}, 2017.

\bibitem{roweis2000nonlinear}
S.~T. Roweis and L.~K. Saul, ``Nonlinear dimensionality reduction by locally
  linear embedding,'' \emph{science}, vol. 290, no. 5500, pp. 2323--2326, 2000.

\bibitem{zhang2003nonlinear}
Z.~Zhang and H.~Zha, ``Nonlinear dimension reduction via local tangent space
  alignment,'' in \emph{International Conference on Intelligent Data
  Engineering and Automated Learning}.\hskip 1em plus 0.5em minus 0.4em\relax
  Springer, 2003, pp. 477--481.

\bibitem{tenenbaum2000global}
J.~B. Tenenbaum, V.~De~Silva, and J.~C. Langford, ``A global geometric
  framework for nonlinear dimensionality reduction,'' \emph{science}, vol. 290,
  no. 5500, pp. 2319--2323, 2000.

\bibitem{weinberger2006introduction}
K.~Q. Weinberger and L.~K. Saul, ``An introduction to nonlinear dimensionality
  reduction by maximum variance unfolding,'' in \emph{AAAI}, vol.~6, 2006, pp.
  1683--1686.

\bibitem{he2004locality}
X.~He and P.~Niyogi, ``Locality preserving projections,'' in \emph{Advances in
  neural information processing systems}, 2004, pp. 153--160.

\bibitem{he2005neighborhood}
X.~He, D.~Cai, S.~Yan, and H.-J. Zhang, ``Neighborhood preserving embedding,''
  in \emph{Tenth IEEE International Conference on Computer Vision (ICCV'05)
  Volume 1}, vol.~2.\hskip 1em plus 0.5em minus 0.4em\relax IEEE, 2005, pp.
  1208--1213.

\bibitem{kokiopoulou2005orthogonal}
E.~Kokiopoulou and Y.~Saad, ``Orthogonal neighborhood preserving projections,''
  in \emph{Fifth IEEE International Conference on Data Mining (ICDM'05)}.\hskip
  1em plus 0.5em minus 0.4em\relax IEEE, 2005, pp. 8--pp.

\bibitem{salem2017manifold}
R.~Salem, ``A manifold learning framework for reducing high-dimensional big
  text data,'' in \emph{2017 12th International Conference on Computer
  Engineering and Systems (ICCES)}.\hskip 1em plus 0.5em minus 0.4em\relax
  IEEE, 2017, pp. 347--352.

\bibitem{wang2018flexible}
W.~Wang, Y.~Yan, F.~Nie, S.~Yan, and N.~Sebe, ``Flexible manifold learning with
  optimal graph for image and video representation,'' \emph{IEEE Transactions
  on Image Processing}, vol.~27, no.~6, pp. 2664--2675, 2018.

\bibitem{pedronette2018unsupervised}
D.~C.~G. Pedronette, F.~M.~F. Gon{\c{c}}alves, and I.~R. Guilherme,
  ``Unsupervised manifold learning through reciprocal knn graph and connected
  components for image retrieval tasks,'' \emph{Pattern Recognition}, vol.~75,
  pp. 161--174, 2018.

\bibitem{pedronette2016correlation}
D.~C.~G. Pedronette and R.~d.~S. Torres, ``A correlation graph approach for
  unsupervised manifold learning in image retrieval tasks,''
  \emph{Neurocomputing}, vol. 208, pp. 66--79, 2016.

\bibitem{arevalo2018manifold}
F.~L.~G. Ar{\'e}valo \emph{et~al.}, ``Manifold learning for spatial audio
  rendering,'' Ph.D. dissertation, Universidade Estadual de Campinas, Faculdade
  de Engenharia El{\'e}trica e de~…, 2018.

\bibitem{yang2016online}
L.~Yang and X.~Wang, ``Online appearance manifold learning for video
  classification and clustering,'' in \emph{International Conference on
  Computational Science and Its Applications}.\hskip 1em plus 0.5em minus
  0.4em\relax Springer, 2016, pp. 551--561.

\bibitem{donoho2003hessian}
D.~L. Donoho and C.~Grimes, ``Hessian eigenmaps: Locally linear embedding
  techniques for high-dimensional data,'' \emph{Proceedings of the National
  Academy of Sciences}, vol. 100, no.~10, pp. 5591--5596, 2003.

\bibitem{zhang2007mlle}
Z.~Zhang and J.~Wang, ``Mlle: Modified locally linear embedding using multiple
  weights,'' in \emph{Advances in neural information processing systems}, 2007,
  pp. 1593--1600.

\bibitem{xiang2011regression}
S.~Xiang, F.~Nie, C.~Pan, and C.~Zhang, ``Regression reformulations of lle and
  ltsa with locally linear transformation,'' \emph{IEEE Transactions on
  Systems, Man, and Cybernetics, Part B (Cybernetics)}, vol.~41, no.~5, pp.
  1250--1262, 2011.

\bibitem{MDS}
J.~B. Kruskal and M.~Wish, \emph{Multidimensional scaling}.\hskip 1em plus
  0.5em minus 0.4em\relax Sage, 1978, vol.~11.

\bibitem{RML}
T.~Lin and H.~Zha, ``Riemannian manifold learning,'' \emph{IEEE Transactions on
  Pattern Analysis and Machine Intelligence}, vol.~30, no.~5, pp. 796--809,
  2008.

\bibitem{LLC}
S.~T. Roweis, L.~K. Saul, and G.~E. Hinton, ``Global coordination of local
  linear models,'' in \emph{Advances in neural information processing systems},
  2002, pp. 889--896.

\bibitem{LE}
M.~Belkin and P.~Niyogi, ``Laplacian eigenmaps for dimensionality reduction and
  data representation,'' \emph{Neural Computation}, vol.~15, pp. 1373--1396,
  2002.

\bibitem{ONPP}
E.~Kokiopoulou and Y.~Saad, ``Orthogonal neighborhood preserving projections,''
  in \emph{Fifth IEEE International Conference on Data Mining (ICDM'05)}.\hskip
  1em plus 0.5em minus 0.4em\relax IEEE, 2005, pp. 8--pp.

\bibitem{SPP}
L.~Qiao, S.~Chen, and X.~Tan, ``Sparsity preserving projections with
  applications to face recognition,'' \emph{Pattern Recognition}, vol.~43,
  no.~1, pp. 331--341, 2010.

\bibitem{gui2012discriminant}
J.~Gui, Z.~Sun, W.~Jia, R.~Hu, Y.~Lei, and S.~Ji, ``Discriminant sparse
  neighborhood preserving embedding for face recognition,'' \emph{Pattern
  Recognition}, vol.~45, no.~8, pp. 2884--2893, 2012.

\bibitem{saul2000introduction}
L.~K. Saul and S.~T. Roweis, ``An introduction to locally linear embedding,''
  \emph{unpublished. Available at: http://www. cs. toronto. edu/\~{}
  roweis/lle/publications. html}, 2000.

\bibitem{van2009dimensionality}
L.~Van Der~Maaten, E.~Postma, and J.~Van~den Herik, ``Dimensionality reduction:
  a comparative,'' \emph{J Mach Learn Res}, vol.~10, no. 66-71, p.~13, 2009.

\end{thebibliography}
\bibliographystyle{IEEEtran}
 
\end{document}